\pdfoutput=1

\documentclass[11pt]{article}

\usepackage{acl} 

\usepackage{times}
\usepackage{latexsym}

\usepackage[T1]{fontenc}

\usepackage[utf8]{inputenc}

\usepackage{microtype}
\usepackage{caption}
\usepackage{graphicx}
\usepackage{makecell}
\usepackage{multirow}
\usepackage{graphicx}
\usepackage{float}
\title{AMS\_ADRN at SemEval-2022 Task 5:  A Suitable Image-text Multimodal Joint Modeling Method for Multi-task Misogyny Identification}

\author{Da Li \\
  Tencent. Inc\\Beijing, China \\
  \texttt{danielsli@tencent.com} \\\And
  Ming Yi \\
  Tencent. Inc\\Beijing, China \\
  \texttt{mingyyi@tencent.com} \\\And
  Yukai He \\
  Tencent. Inc\\Beijing, China \\
  \texttt{yukaihe@tencent.com}
}

\begin{document}
\maketitle
\begin{abstract}
Women are influential online, especially in image-based social media such as Twitter and Instagram. However, many in the network environment contain gender discrimination and aggressive information, which magnify gender stereotypes and gender inequality. Therefore, the filtering of illegal content such as gender discrimination is essential to maintain a healthy social network environment.
In this paper, we describe the system developed by our team for SemEval-2022
Task 5: Multimedia Automatic Misogyny Identification. More specifically, we introduce two novel systems to analyze these posts: a multimodal multi-task learning architecture that combines Bertweet for text encoding with ResNet-18 \cite{he2016deep} for image representation, and a single-flow transformer structure which combines text embeddings from BERT-Embeddings and image embeddings from several different modules such as EfficientNet \cite{tan2019efficientnet} and ResNet \cite{he2016deep}. In this manner, we show that the information behind them can be properly revealed. Our approach achieves good performance on each of the two subtasks of the current competition, ranking 12th for Subtask A (0.746 macro F1-score), 10th for Subtask B (0.706 macro F1-score) while exceeding the official baseline results by high margins.
\end{abstract}

\section{Introduction}

Women are influential online, especially in image-based social media such as Twitter and Instagram : 78 percent of women use social media multiple times a day, compared with 65 percent of men.\cite{MAMI2022} However, although the Internet has opened new opportunities for women, systematic offline inequality and discrimination are replicated in cyberspace in the form of offensive content against them. The popular communication tool in social media platforms is MEME. Memetics is essentially an image, which is characterized by the content of the image and the poster cover text introduced by human beings. Its main goal is fun or irony. Although most of them were created for a joke, in a very short period of time, people began to use them as a form of hatred against women, and then magnified gender stereotypes and gender inequality. Women are exposed to gender discrimination and aggressive information in the network environment. Therefore, the filtering of illegal content such as gender discrimination is essential to maintain a healthy social network environment.

In this paper, we describe the system developed by our team for SemEval-2022 Task 5: Multimedia Automatic Misogyny Identification. More specifically, we introduce two novel systems to analyze these posts: a multimodal multi-task learning architecture that combines Bertweet \cite{bertweet} for text encoding with ResNet-18 \cite{he2016deep} for image representation, and a single-flow transformer structure which combines text embeddings from BERT-Embeddings and image embeddings from several different modules such as EfficientNet \cite{tan2019efficientnet} and ResNet \cite{he2016deep}.

\section{Related Work}

{\bf Language Model Pre-training.}
The MLM-based pre-training method used in BERT \cite{BERT} opens up the pre-training paradigm of language model based on Transformer structure. RoBERTa \cite{liu2019roberta} carefully measured the influence of key hyperparameters and training data size, and further enhanced the effect. SpanBERT \cite{joshi2020spanbert} extends BERT by masking continuous random spans rather than random markers, and trains the span boundary representation to predict the whole content of the shielded span without relying on a single marker representation. MacBERT \cite{cui2020revisiting} improved RoBERTa in several aspects, especially using MLM as the correction masking strategy (Mac).  BERTweet \cite{bertweet} is the first large-scale language model for English Tweets and produced better performance results than the previous state-of-the-art models on
three Tweet NLP tasks: Part-of-speech tagging, Named-entity recognition and text classification. 

{\bf Visual and Language Multimodal Learning.} 
Some studies explored cross-modal transmission between images and text. Some ideas are related to cross-model representation learning \cite {C14,C15}, aiming to generate representations that effectively correlate different sensor modes. Recently, several ideas have been proposed to focus on the transformer backbone by improving the pre-training strategy rather than optimizing the backbone structure. VideoBERT \cite{C24} learns the bidirectional joint distribution of visual and linguistic tag sequences, which are respectively from the vector quantization of frame data and the ready-made speech recognition output. VisualBERT \cite {Visualbert} uses MLM to pre-train specific tasks, which uses the same type ID for language \& visual input. VL-BERT \cite{C17} uses a simple Transformer model as the backbone, and uses a mixture of region of interest ( ROI ) and words in the image as input. 

In addition, some ideas have become focused on improving the transformer backbone network. Unicoder-VL \cite{C18} uses three pre-training tasks, including mask language modeling (MLM), mask object classification (MOC) and visual language matching (VLM), to learn the joint representation of vision and language. ViLBERT \cite{C9} extends the BERT architecture to a multi-modal dual-flow model and processes it in separate flows interacting through the common attention converter layer

\section{Task Description}

The proposed task, i.e. Multimedia Automatic Misogyny Identification (MAMI) consists in the identification of misogynous memes, taking advantage of both text and images available as source of information. The task will be organized around two main sub-tasks:
\begin{itemize}
	\item Sub-task A: a basic task about misogynous meme identification, where a meme should be categorized either as misogynous or not misogynous;
	\item Sub-task B: an advanced task, where the type of misogyny should be recognized among potential overlapping categories such as stereotype, shaming, objectification and violence.
\end{itemize}

\section{Methodology}

For this task, we have tried a variety of modeling and optimization methods, which are described in detail as follows.

\begin{figure*}[htb]
	\begin{minipage}[b]{1.0\linewidth}
		\centering
		\centerline{\includegraphics[width=16cm]{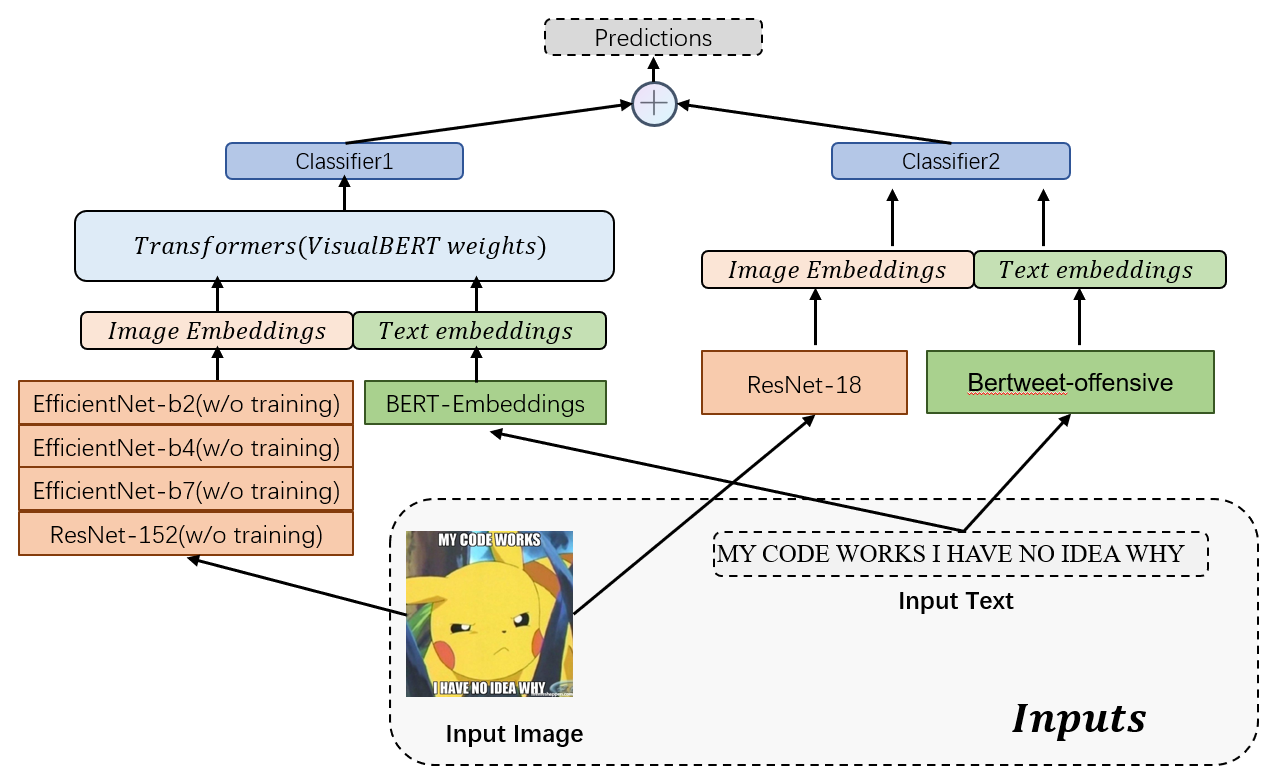}}
	\end{minipage}
	\caption{An overview of the two model structures. Transformer single-flow structure on the left and double-tower structure on the right.}
	\label{fig:fig1}
\end{figure*}

\subsection{Model design}

We designed two kinds of multimodal model schemes with different architectures : transformer single-flow structure and double-tower structure.

{\bf Transformer Single-flow Structure.} The model of this structure only contains a set of longitudinal stacks of transformer modules. For image and text features, after their respective processing and stitching, together as the input of the model, so called single-stream transformer model. The advantage of this model is that it can express image and text cross-modal features in the same vector space, which is equivalent to mapping multi-modal features to single-modal vector representation space. The prediction effect of single-flow model is very dependent on pre-training. This work uses VisualBERT \cite{Visualbert} pretraining weights on COCO-VQA tasks. The text features directly use BERT-base-uncased \cite{BERT} tokenizer for word segmentation and post-vectorization. The image features are vectorized by ResNet-152 \cite{he2016deep}, EfficientNet-b2, EfficientNet-b4 and EfficientNet-b7 models \cite{tan2019efficientnet} to improve the richness of image feature expression. Finally, the image and the text vector are spliced together as the input of the model.

{\bf Double-tower Structure.} Different from single-flow models, the double-tower model is actually two parallel models. For images and text, each builds a model structure and adds MLP to the output side to parallel them for joint training. Although the twin-tower structure cannot retain a large amount of prior knowledge through pre-training as the single-flow model does, nor can it fully excavate the cross-modal interaction characteristics as the twin-flow model does, the twin-tower structure is less dependent on the amount of training sample data, and is more flexible. Compared with the single-flow and double-flow transformer structures, it is more suitable for this task. Due to less training data, in order to avoid overfitting, we choose a relatively simple network structure, image side feature representation model using ResNet-18 \cite{he2016deep}, text side feature representation model using Bertweet.

{\bf Model Ensemble.} In order to make full use of the differences between models to achieve the best prediction effect, we use the weighted average method to obtain the final prediction results. The method is as follows:
\begin{eqnarray}
	Y_{pred}=\alpha \cdot Y_1 + (1 - \alpha) \cdot Y_2
\end{eqnarray}
Where $Y_{pred}$ is the final prediction probability, $Y_1$ is the single flow model prediction, $Y_2$ is the double tower model prediction, $\alpha=0.1$.

\subsection{Training Framework}

\begin{figure*}[htb]
	\begin{minipage}[b]{1.0\linewidth}
		\centering
		\centerline{\includegraphics[width=12cm]{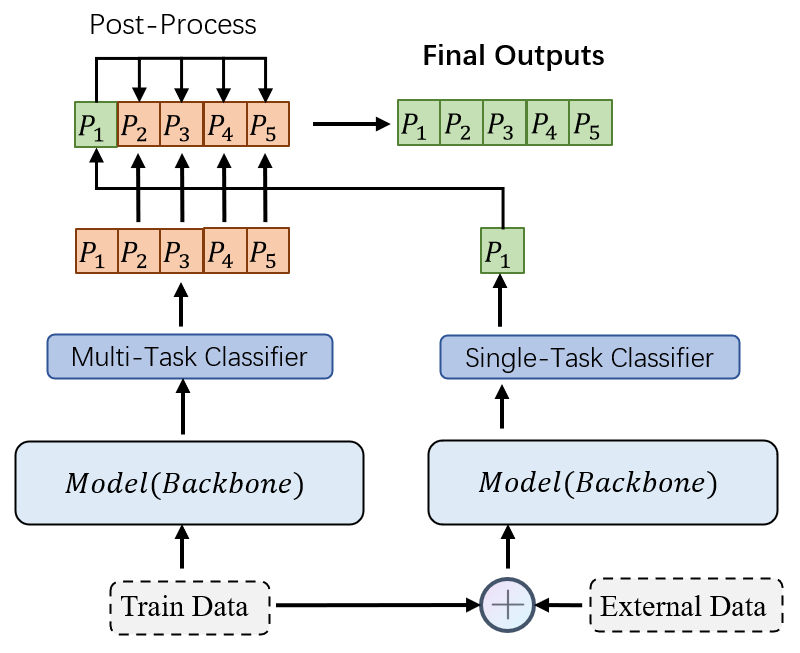}}
	\end{minipage}
	\caption{An overview of the proposed three-stage training framework. Firstly, we only use the training data to train the multi-task model and obtain the prediction probability of each task. Then, we use the external data to expand the training data to train the single task model and obtain the probability of 'misogynous'. Finally, we use the probability of 'misogynous' to post-process other tasks and obtain the final prediction results.}
	\label{fig:fig2}
\end{figure*}

As shown in Fig \ref{fig:fig2}, we use a three-stage training framework consists of multi-task binary classification for sub-task B, single-task binary classification for sub-task A and post-processing.

{\bf Stage1: Multi-task binary classification.}
Sub-task A is a single-task binary classification problem, and sub-Task B is a multi-task binary classification problem. We tried to define sub-Task B as a single-task binary classification, so that the amount of data becomes five times that of multi-task binary classification to expand the training samples. However, the experimental results show that the effect is basically the same as that of multi-task binary classification. Therefore, we first define a multi-task binary task and obtain the respective prediction probabilities of the model for misogynous, stereotype, shaming, objectification and violence.

{\bf Stage2: Single-task binary classification.}
For sub-task A, we use external data from the memotion task \cite{chhavi2020memotion,Memotion2} for the misogynous binary classification. Although these data predict different targets, they are homologous to the task data. Through the analysis, it is found that there is almost no gender discrimination in the samples of 'Not Offensive' and 'Slight Offensive'. Therefore, this part of the external data is used to expand the training set, which greatly improves the prediction results of sub-task A. The reason why external data is not used in multi-task stage is that the proportion of positive samples in 'shaming', 'stereotype', 'objectification' and 'violence' tasks is very low, and increasing the number of negative samples will increase the prediction difficulty of these fields, so only external data is used in this stage.

{\bf Stage3: Post-processing.}
Through analysis, it is found that when 'shaming', 'stereotype', 'objectification' and 'violence' are positive samples, 'misogynous' must be positive samples. Due to the higher F1 value predicted by 'misogynous' in the second stage, the prediction results of 'shaming', 'stereotype', 'objectification' and 'violence' are corrected by the prediction results of 'misogynous' in the second stage. We modify the prediction results of samples 'shaming', 'stereotype', 'objectification' and 'violence' with 'misogynous' prediction 0 to 0, which will improve the prediction results.

%
%
%

\section{Experiments}

This section is organized as follows. First we introduce the MAMI dataset \cite{MAMI2022}, then we introduce the experimental setup in detail. Finally, we show the effectiveness of our proposed method on MAMI datasets.

\subsection{Datasets Analysis}

The number of samples in the dataset is shown in table \ref{tab:tab1}. Sub-task A uses external data, so the training data contains 10000 original samples and 14820 external data samples, and the predicted data contains 1000 samples; sub-task B does not use external data, so the training samples contain only 10000 original items and the predicted data also contains 1000 samples.

In the original 10000 training samples, there are five prediction labels, 'misogynous', 'shaming', 'stereotype', 'objectification' and 'violence'. Their respective number and proportion of positive samples are shown in table \ref{tab:tab2}.
\begin{table}
	\centering
	\begin{tabular}{lrr}
		\hline
		\textbf{Dataset} & \textbf{Sub-task A} & \textbf{Sub-task B}\\
		\hline
		train & 10000 & 10000 \\
		external-train & 14820 & 0 \\
		test & 1000 & 1000 \\\hline
	\end{tabular}
	\caption{Datasets statistics.}
	\label{tab:tab1}
\end{table}

\begin{table}
	\centering
	\begin{tabular}{lrr}
		\hline
		\textbf{Category} & \textbf{Samples} & \textbf{Percentage}\\
		\hline
		misogynous & 5000 & 50.0\% \\
		shaming & 1274 & 12.7\% \\
		stereotype & 2810 & 28.1\% \\
		objectification & 2202 & 22.0\% \\
		violence & 953 & 9.5\% \\\hline
	\end{tabular}
	\caption{Datasets analysis.}
	\label{tab:tab2}
\end{table}

\subsection{Experiment Settings}

We implement our model using Pytorch. Using a workstation with an Intel Xeon processor, 64GB of RAM and a Nvidia P40 GPU for training. We apply AdamW as an optimization algorithm with $10\%$ steps of warmup. For the hyperparameters, we set epochs=$10$, batch size=$64$. We also set the earlystop patience epoch as $3$. We resize all images to $256\times256$ RGB pixels, and set the maximum cut length of text to 64. For the single-flow model, learning rate is set to $5\times10^{-5}$. For the double-tower model, we use the sub-regional learning rate strategy. The learning rate of text module is $5\times10^{-5}$, the learning rate of image module is $1\times10^{-4}$, and the learning rate of combined multi-layer percetron (MLP) module is $1\times10^{-3}$.
We used multi-label stratified k-fold method to split training data into 5 folds. 

Augmentation method used in image preprocessing. In the field of computer vision, because the image data have high dimensional characteristics, it is necessary to expand the training data, and usually do so. We use the following procedures to expand. In the training phase, we use (i) random resizing and cropping, (ii) random horizontal flipping, and (iii) random vertical flipping. In the inference phase, we use “five-crop inference” for robust prediction. This is essentially an average ensemble of the
predictions on augmented images.

\subsection{Experimental Results}

Table \ref{tab:tab3} contains the results obtained from running the experiments. It can be seen that the prediction effect of single-flow model is worse than that of double-tower model, because the prediction effect of single-flow model is very dependent on pre-training. For the pre-training of single-stream model, most of the current mainstream models such as VisualBERT, UNITER \cite{chen2019uniter} and so on are based on ImageNet \cite{C27}, COCO \cite{lin2014microsoft} and other data sets for graph-text matching pre-training. There are some differences between these data sets and the scene of this task : (i) Twitter images contain text, and it is essential for emotional tendency expression, while ImageNet, COCO and other data sets contain almost no text ; (ii) The text of ImageNet, COCO and other data sets is the semantic expression of the image, and the text does not appear directly in the image, but the text information in the twitter sample is the text extracted from the image by OCR, which can be said to be a subset of image features from the feature information. The pre-training method of image-text matching is not very suitable. These reasons limit the prediction effect of the single-flow transformer structure model to a certain extent. Although compared with baseline, the potential of the model is still quite large. Sufficient training data and appropriate pre-training task design can further significantly improve the prediction results.In the single model, the twin-tower model has the best prediction results. Precisely because a particular module of the model contains only prior knowledge of a particular modality, it is more adaptable to different scenes. The cross-modal features are only learned through the current task, and do not depend on the prior knowledge. When the training data are not sufficient for pre-training, this modeling method can achieve the best results.

Benefit from the huge difference in the feature representation of the two models, the difference of prediction results between single-flow model and double-tower model is very large, and better prediction results can be obtained by simple weighted average method. Finally, the model obtained by subtask A is used to post-process other prediction results, and the final best result is obtained.
\begin{table}
	\setlength{\tabcolsep}{0.6mm}{
	\centering
	\begin{tabular}{lrr}
		\hline
		\textbf{Method} & \textbf{Sub-task A} & \textbf{Sub-task B}\\
		\hline
		organizers baseline & 0.650 & 0.621 \\
		transformer single-flow & 0.658 & 0.685 \\
		double-tower & 0.684 & 0.706 \\
		model ensemble & 0.682 & 0.707\\
		post-processing & \textbf{0.746} & \textbf{0.708} \\\hline
	\end{tabular}}
	\caption{Results on MAMI dataset.}
	\label{tab:tab3}
\end{table}

\subsection{Error Analysis}

\begin{figure*}[htbp]
	\vspace{-2.6cm}
	\begin{minipage}[b]{1.0\linewidth}
		\centering
		\centerline{\includegraphics[width=16cm]{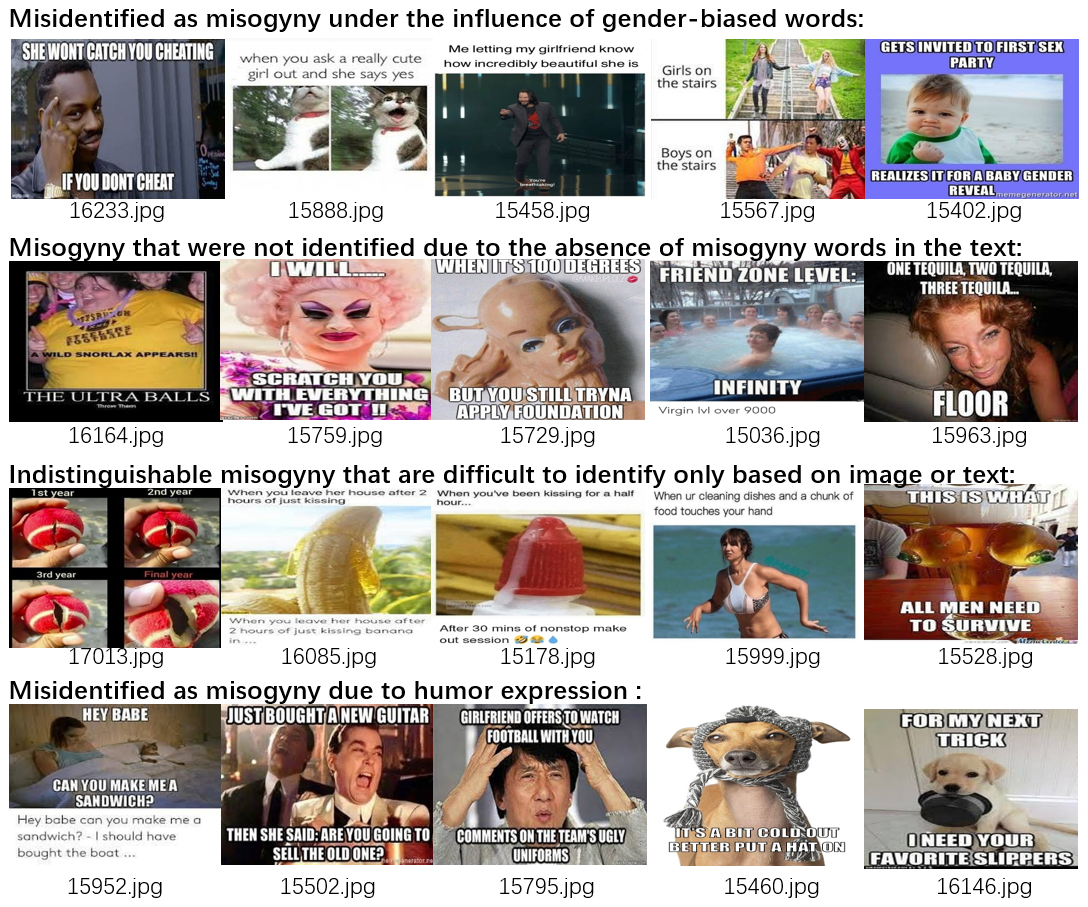}}
		\vspace{-0.3cm}
	\end{minipage}
	\caption{Error analysis of typical samples.}
	\label{fig:fig3}
	\vspace{-2cm}
\end{figure*}

As shown in Table \ref{tab:tab4} and Fig \ref{fig:fig3}, we select some typical samples that can reflect the performance of the methods for analysis. In Table \ref{tab:tab4}, the first two columns are the file name and the text recognized in the image. SF, DT, ME and PP represent the prediction results of transformer single-flow, double-tower, model ensemble and post-processing methods in Table \ref{tab:tab2} on the misogyny task, respectively.

We can see that the transformer single-flow method is more dependent on text information than image. For some non-misogyny images, due to some gender-biased words like "she", "cute girl" and "sex", the transformer single-stream method mispredicts misogyny images, and the double-tower method can correctly distinguish them. For some misogyny images, the single-flow method fails to predict correctly because there is no misogyny tendency in the text.

\begin{table*}[htbp]
	\small
	\vspace{-2cm}
	\setlength{\tabcolsep}{1mm}{
		\centering
		\begin{tabular}{llllll}
			\hline
			\textbf{file name} & \textbf{Text Transcription} & \textbf{SF} & \textbf{DT} & \textbf{ME} & \textbf{PP}\\
			\hline
			\multicolumn{6}{l}{\textbf{\emph{Misidentified as misogyny under the influence of gender-biased words}}} \\
			16233.jpg & \makecell[l]{SHE WONT CATCH YOU CHEATING ingtip.com IF YOU DONT CHEAT Opening \\Men Tut-Thur Fri-Sal Sunday} & 1 & 0 & 0 & 0 \\
			15888.jpg & when you ask a really cute girl out and she says yes & 1 & 0 & 0 & 0 \\
			15458.jpg & Me letting my girlfriend know how incredibly beautiful she is You're breathtaking! & 1 & 0 & 0 & 0 \\
			15567.jpg & Girls on the stairs Boys on the stairs & 1 & 0 & 0 & 0 \\
			15402.jpg & \makecell[l]{GETS INVITED TO FIRST SEX PARTY 36 REALIZES IT FOR A BABY GENDER \\REVEAL meme memegenerator.net} & 1 & 0 & 0 & 0 \\
			\hline
			\multicolumn{6}{l}{\textbf{\emph{Misogyny samples that were not identified due to the absence of misogyny words in the text}}} \\
			16164.jpg & \makecell[l]{TSBURGH Bat TEELERS DOTBALL \$65 A WILD SNORLAX APPEARS!! THE \\ULTRA BALLS Throw Them} & 0 & 1 & 1 & 1 \\
			15759.jpg & IWILL... SCRATCH YOU WITH EVERYTHING I'VE GOT !! & 0 & 1 & 1 & 1 \\
			15729.jpg & \makecell[l]{WHEN IT'S 100 DEGREES @MAKEUPLOLZ BUT YOU STILL TRYNA APPLY \\FOUNDATION} & 0 & 1 & 1 & 1 \\
			15036.jpg & FRIEND ZONE LEVEL: INFINITY Virgin lvl over 9000 & 0 & 1 & 1 & 1 \\
			15963.jpg & ONE TEQUILA, TWO TEQUILA, THREE TEQUILA... FLOOR & 0 & 1 & 1 & 1 \\
			\hline
			\multicolumn{6}{l}{\textbf{\emph{Indistinguishable misogyny samples that are difficult to identify only based on image or text}}} \\
			17013.jpg & 1st year 3rd year 2nd year Final year & 0 & 1 & 0 & 1 \\
			16085.jpg & \makecell[l]{When you leave her house after 2 hours of just kissing When you leave her house after 2 \\hours of just kissing banana in ...} & 0 & 0 & 0 & 1 \\
			15178.jpg & \makecell[l]{When you've been kissing for a half hour... Via MehsilyPre h.com After 30 mins of \\nonstop make out session} & 0 & 0 & 0 & 1 \\
			15999.jpg & When ur cleaning dishes and a chunk of food touches your hand @MARVY & 1 & 0 & 0 & 1 \\
			15528.jpg & CABE THIS IS WHAT ALL MEN NEED TO SURVIVE Meme Center & 0 & 1 & 0 & 1 \\
			\hline
			\multicolumn{6}{l}{\textbf{\emph{Misidentified as a misogyny sample because of humor}}} \\
			15952.jpg & \makecell[l]{HEY BABE CAN YOU MAKE ME A SANDWICH? Hey babe can you make me a \\sandwich? I should have bought the boat...} & 1 & 1 & 1 & 0 \\
			15502.jpg & \makecell[l]{JUST BOUGHT A NEW GUITAR THEN SHE SAID: ARE YOU GOING TO SELL \\THE OLD ONE? hengenerator.net} & 1 & 1 & 1 & 0 \\
			15795.jpg & \makecell[l]{GIRLFRIEND OFFERS TO WATCH FOOTBALL WITH YOU COMMENTS ON THE \\TEAM'S UGLY UNIFORMS} & 1 & 1 & 1 & 0 \\
			15460.jpg & IT'S A BIT COLD OUT BETTER PUT A HAT ON & 1 & 1 & 1 & 0 \\
			16146.jpg & FOR MY NEXT TRICK ONEED YOUR FAVORITE SLIPPERS & 0 & 1 & 1 & 0 \\
			\hline
	\end{tabular}}
	\vspace{-0.2cm}
	\caption{Error analysis of typical samples.}
	\label{tab:tab4}
\end{table*}

Many samples of misogyny are very obscure and do not tend to discriminate either from text or from images alone, but there is a combination of discriminatory hints that post-processing method is more capable of detecting such samples. At the same time, the post-processing method also calibrated some samples that were mistakenly identified as misogyny due to humor expression.

\section{Conclusions and Future Works}

In this paper, we describe the system developed by our team for SemEval-2022 Task 5: Multimedia Automatic Misogyny Identification. More specifically, we introduce two novel systems to analyze these posts: a multimodal multi-task learning architecture that combines Bertweet for text encoding with ResNet-18 \cite{he2016deep} for image representation, and a single-flow transformer structure which combines text embeddings from BERT-Embeddings and image embeddings from several different modules such as EfficientNet \cite{tan2019efficientnet} and ResNet \cite{he2016deep}.In addition, we also use the model fusion and the post-processing method of prediction results, and improve the prediction results. 

In the future, we will try to get more unlabeled sample data to pre-train the transformer single-flow method more fully and look forward to getting better multimodal methods in Twitter scenarios.

\small
\bibliography{anthology,custom}

\end{document}